\documentclass[letterpaper, 10 pt, journal, twoside]{ieeetran}

\usepackage{graphicx}
\usepackage{amsmath}
\usepackage{amssymb}
\usepackage{subcaption}
\usepackage{array}
\usepackage{flushend}
\usepackage{cite}
\usepackage{hyperref}
\usepackage{float}
\usepackage{microtype}

\hypersetup{
    colorlinks=true,
    linkcolor=black,
    urlcolor=black,
    citecolor=black
}

\usepackage{amsmath, amssymb}
\usepackage{extraipa}

\usepackage{xifthen}
\usepackage{color}

\newcommand{\ba}{\begin{eqnarray}}
\newcommand{\ea}{\end{eqnarray}}

\newcommand{\unit}[1]{\mbox{$\mathrm{\,#1}$}}

\renewcommand{\deg}{\mbox{${}^\circ$}}

\newcommand{\R}{\mathbb{R}}

\newcommand{\presup}[1]{\,{}^{\scriptscriptstyle #1}\!}

\newcommand{\pose}[1][ZZZZ]{\ifthenelse{\equal{#1}{ZZZZ}}{}{\presup{#1}}{\mathbf{\xi}}}
\newcommand{\estpose}[1][ZZZZ]{\ifthenelse{\equal{#1}{ZZZZ}}{}{\presup{#1}}{\mathbf{\hat{\xi}}}}
\newcommand{\hpose}[1][ZZZZ]{\ifthenelse{\equal{#1}{ZZZZ}}{}{\presup{#1}}{\hat{\mathbf{\xi}}}}
\newcommand{\posedot}[1][ZZZZ]{\ifthenelse{\equal{#1}{ZZZZ}}{}{\presup{#1}}{\mathbf{\nu}}}

\newcommand{\q}[1][ZZZZ]{\ifthenelse{\equal{#1}{ZZZZ}}{}{\presup{#1}}{\mathring{q}}}

\DeclareMathAlphabet{\mathitbf}{OML}{cmm}{b}{it}
\newcommand{\twist}[2][ZZZZ]{\ifthenelse{\equal{#1}{ZZZZ}}{}{\presup{#1}}{\mathcal{S}}}
\renewcommand{\vec}[2][ZZZZ]{\ifthenelse{\equal{#1}{ZZZZ}}{}{\presup{#1}}{\mathitbf{#2}}}

\newcommand{\hvec}[2][ZZZZ]{\ifthenelse{\equal{#1}{ZZZZ}}{}{\presup{#1}}{\tilde{\vec{#2}}}}
\newcommand{\obvec}[2][ZZZZ]{\ifthenelse{\equal{#1}{ZZZZ}}{}{\presup{#1}}\rlap{${\overbridge{\phantom{$\vec{#2}$}}}$}\vec{#2}}
\newcommand{\evec}[2][ZZZZ]{\ifthenelse{\equal{#1}{ZZZZ}}{}{\presup{#1}}{\hat{\vec{#2}}}}
\newcommand{\bvec}[2][ZZZZ]{\ifthenelse{\equal{#1}{ZZZZ}}{}{\presup{#1}}{\bar{\vec{#2}}}}

\newcommand{\dvec}[2][ZZZZ]{\ifthenelse{\equal{#1}{ZZZZ}}{}{\presup{#1}}{\dot{\vec{#2}}}}
\newcommand{\ddvec}[2][ZZZZ]{\ifthenelse{\equal{#1}{ZZZZ}}{}{\presup{#1}}{\ddot{\vec{#2}}}}

\newcommand{\mat}[2][ZZZZ]{\ifthenelse{\equal{#1}{ZZZZ}}{}{\presup{#1}\,}{{\boldsymbol #2}}}
\newcommand{\dmat}[2][ZZZZ]{\ifthenelse{\equal{#1}{ZZZZ}}{}{\presup{#1}\,}{{\dot{\boldsymbol #2}}}}
\newcommand{\emat}[2][ZZZZ]{\ifthenelse{\equal{#1}{ZZZZ}}{}{\presup{#1}\,}{\hat{\boldsymbol#2}}}
\newcommand{\matfn}[3][ZZZZ]{\ifthenelse{\equal{#1}{ZZZZ}}{}{\presup{#1}}{{\mat{#2}}\left(#3\right)}}
\newcommand{\Rt}[2][ZZZZ]{\ifthenelse{\equal{#1}{ZZZZ}}{}{\presup{#1}}{{\bf R}\left(#2\right)}}

\newcommand{\point}[2][ZZZZ]{\ifthenelse{\equal{#1}{ZZZZ}}{}{\presup{#1}}{\mathbf{\mathrm{#2}}}}

\newfont{\School}{pncr}
\newfont{\eightTR}{pncr at 8pt}

\usepackage{color}
\usepackage{fancyvrb}
\fvset{formatcom=\color{blue},fontseries=c,fontfamily=courier,xleftmargin=4mm,commentchar=!}

\DefineVerbatimEnvironment{Code}{Verbatim}{formatcom=\color{blue},fontseries=c,fontfamily=courier,fontsize=\footnotesize,xleftmargin=4mm,commentchar=!}

\DefineVerbatimEnvironment{CodeSmall}{Verbatim}{formatcom=\color{blue},fontseries=c,fontfamily=courier,fontsize=\scriptsize,xleftmargin=1mm,commentchar=!}

\DefineVerbatimEnvironment{CodeNum}{Verbatim}{numbers=left,numbersep=4pt,formatcom=\color{blue},fontseries=c,fontfamily=courier,fontsize=\footnotesize,xleftmargin=4mm}

\newcommand{\model}[1]{\index{code}{#1@\textit{#1}}\ifthenelse{\boolean{draft}}{{\color{green}\Verb+#1+}}{\Verb+#1+}}
\newcommand{\block}[1]{\ifthenelse{\boolean{draft}}{{\color{green}\Verb+#1+}}{\textsf{#1}}}

\newcommand{\func}[2][ZZZZ]{\ifthenelse{\equal{#1}{ZZZZ}}{\index{code}{#2}}{\index{code}{#1}}\ifthenelse{\boolean{draft}}{{\color{green}\Verb+#2+}}{\Verb+#2+}}

\newcommand{\methodb}[2]{\index{code}{#1@\textbf{#1}!.#2}\ifthenelse{\boolean{draft}}{{\color{magenta}\Verb+#1.#2+}}{\Verb+#1.#2+}}
\newcommand{\method}[2]{\index{code}{#1@\textbf{#1}!.#2}\ifthenelse{\boolean{draft}}{{\color{magenta}\Verb+#2+}}{\Verb+#2+}}
\newcommand{\class}[1]{\index{code}{#1@\textbf{#1}}\ifthenelse{\boolean{draft}}{{\color{cyan}\Verb+#1+}}{\Verb+#1+}}
\newcommand{\property}[1]{\index{property}{#1}\ifthenelse{\boolean{draft}}{{\color{cyan}\Verb+#1+}}{\Verb+#1+}}

\newcommand{\SE}[1]{\ensuremath{\mathrm{{\bf SE}(#1)}}}

\begin{document}

\title{
NEO: A Novel Expeditious Optimisation Algorithm for Reactive Motion Control of Manipulators}

\author{Jesse Haviland$^{1}$, Peter Corke$^{1}$%
\thanks{Manuscript received: October, 15, 2020; Revised December, 26, 2020; Accepted January, 6, 2021. This paper was recommended for publication by Editor Clement Gosselin upon evaluation of the Associate Editor and Reviewers' comments. This research was conducted by the Australian Research Council project number CE140100016, and supported by the QUT Centre for Robotics.}%
\thanks{$^{1}$Jesse Haviland and Peter Corke are with the Australian Centre for Robotic Vision (ACRV), Queensland University of Technology Centre for Robotics (QCR), Brisbane, Australia
        {\tt\small j.haviland@qut.edu.au, peter.corke@qut.edu.au}. }%
\thanks{Digital Object Identifier (DOI): 10.1109/LRA.2021.3056060}%
}

\markboth{IEEE Robotics and Automation Letters. Preprint Version. Accepted January, 2021}
{Haviland \MakeLowercase{\textit{et al.}}: NEO: A Novel Expeditious Optimisation Algorithm for Reactive Motion Control of Manipulators} 

\maketitle

\begin{abstract}
We present NEO, a fast and purely reactive motion controller for manipulators which can avoid static and dynamic obstacles while moving to the desired end-effector pose. Additionally, our controller maximises the manipulability of the robot during the trajectory, while avoiding joint position and velocity limits. NEO is wrapped into a strictly convex quadratic programme which, when considering obstacles, joint limits, and manipulability on a 7 degree-of-freedom robot, is generally solved in a few ms. While NEO is not intended to replace state-of-the-art motion planners, our experiments show that it is a viable alternative for scenes with moderate complexity while also being capable of reactive control. For more complex scenes, NEO is better suited as a  reactive local controller, in conjunction with a global motion planner. We compare NEO to motion planners on a standard benchmark in simulation and additionally illustrate and verify its operation on a physical robot in a dynamic environment. We provide an open-source library which implements our controller.
\end{abstract}


\IEEEpeerreviewmaketitle

\section{Introduction}

\IEEEPARstart{T}{he} real world is not static and consequently, manipulators must be up to the challenge of reacting to the unpredictable nature of the environment they work in. Getting a manipulator's end-effector from point A to point B is a fundamental problem in robotic control. While the task is seemingly simple, there can be numerous causes of failure, especially as the environment becomes cluttered and dynamic. Dynamic obstacle avoidance is essential for manipulators to be robust in such environments.

Each iteration of the control loop must be able to consider the state of both environment and robot to guarantee safe and reliable robot operation. The current focus in obstacle avoidance for manipulators has come at the cost of greater up-front computational load, leading to open-loop motion plans.
Planners compute a sequence of joint coordinates which at run time become joint velocities.
In contrast, a reactive approach capable of avoiding non-stationary obstacles works directly with joint velocities.

We consider the differential kinematics and compute a set of joint velocities that \emph{steer} the end-effector towards the goal in task space. This classic and purely reactive approach is known as resolved-rate motion control (RRMC) and is cheap to compute and easily able to run at over 1000 Hz. In this
paper we add additional capabilities to the RRMC approach while maintaining its highly reactive capability.

Differential kinematics also allows us to capture the rate of change of the distance between any part of the robot and any obstacle present in the environment. By exploiting such relationships, a reactive controller can be developed that will avoid colliding with obstacles. The resulting controller may be over-constrained and unable to achieve the goal pose, but we can employ two strategies to resolve this.

Firstly, we could employ a kinematically redundant manipulator with more degrees of freedom than is necessary to reach any pose within its task space -- these are now increasingly common. Secondly, we could relax the path specification to allow for intentional error, called \textit{slack} which allows the end-effector to diverge from the straight-line trajectory to dodge obstacles.

\begin{figure}[t]
    \centering
    \includegraphics[height=4.9cm]{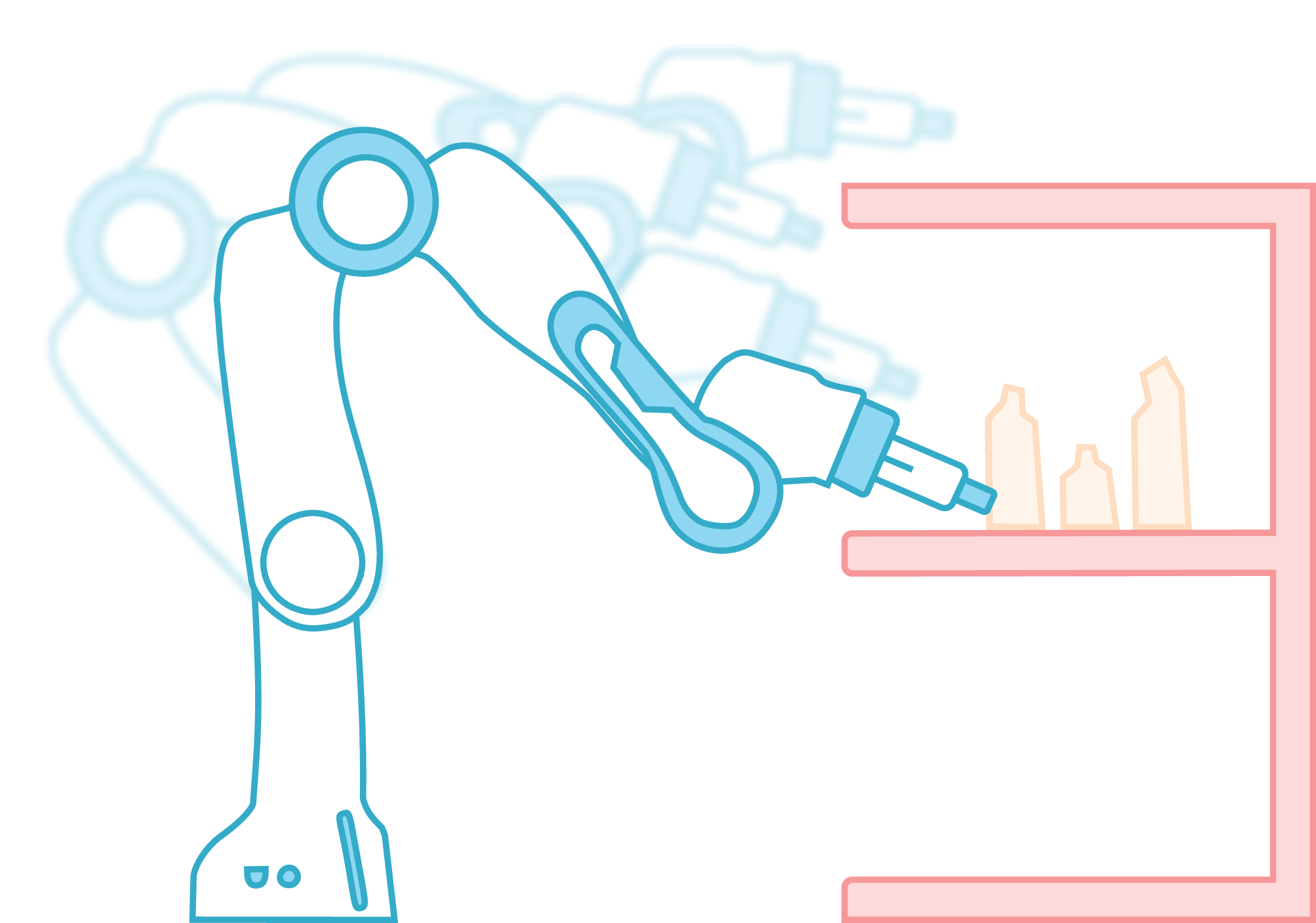}
    \caption{
        Our NEO controller drives a robot's joint velocities such that the robot does not collide with obstacles (that may be non-stationary), does not exceed joint position (translational or angular) or velocity limits and maximises the manipulability of the robot, purely reactively.
    }
    \vspace{-12pt}
    \label{fig:cover}
\end{figure}

To provide the best chance for the robot to be able to react in a volatile environment, we must consider the robot's conditioning. A measure of manipulability, devised in \cite{manip}, describes how well-conditioned a manipulator is to achieve any arbitrary velocity. Therefore, by also maximising the manipulability of the robot, we can decrease the likelihood of robot failure due to singularity while also improving robot obstacle avoidance.

The contributions of this paper are:
\begin{enumerate}
    \item a reactive motion controller for serial-link manipulators (fully actuated or redundant) that can achieve a desired end-effector pose with the ability to dodge stationary and non-stationary obstacles, avoid joint limits and maximise manipulability.  
    \item experimental validation in simulation on a published motion planning benchmark as well as on a physical Franka-Emika Panda robot.
    \item an open-source Python library providing all tools necessary to implements our controller on any serial-link manipulator \cite{rtb}. Benchmark code and implementation details are provided at 
    \href{https://jhavl.github.io/neo}{jhavl.github.io/neo}.
\end{enumerate}

\section{Related Work}\label{sec:related_work}

Classical planning methods will find the optimal path according to the supplied criteria. However, their computation time means they are restricted to use as open-loop planners. Motion planners such as STOMP \cite{stomp}, CHOMP \cite{chomp} and Trajopt \cite{trajopt} use sampling-based methods to solve general high-dimensional motion planning problems not just for manipulators. Motion planners can typically incorporate collision avoidance as well as other constraints such as end-effector orientation throughout the motion. In some cases, motion-planners may need post-processing to remove jerky or extraneous motions due to sampling.

STOMP uses a stochastic and gradient-free scheme to optimise a trajectory while minimising a cost function. The stochastic nature of the scheme means that it can overcome local minima which purely gradient-based approaches like CHOMP and Trajopt can be trapped by. CHOMP and Trajopt both utilise novel methods for solving optimisation problems while incorporating obstacles.

Planning times of these algorithms preclude their use in dynamic environments. 
In one benchmark \cite{trajopt} planning time varies from 0.19 seconds (for Trajopt) to 4.91 seconds (for CHOMP).

Path planners output a sequence of poses for which the inverse kinematic (IK) solution is computed to obtain a sequence of joint coordinates. IK can be solved analytically on most robots, but it is more commonly solved through an optimisation framework \cite{ik0}. The optimisation problem can be augmented, through the cost function, with equality or inequality constraints to provide additional functionality. 
IK solvers can therefore provide joint configurations which are free from collisions and avoid joint limits. Typically only redundant robots with an exploitable null space can achieve useful benefits. 

However, IK alone cannot provide a valid path to the goal. Rather than consider the pose or joint configuration at each time step we can consider velocity. This approach, using differential kinematics, is commonly referred to as Resolved-Rate Motion Control (RRMC) \cite{rrmc}.

RRMC uses the derivative of the forward kinematics to choose joint velocities which will drive the robot in a straight line from the starting end-effector pose to the desired end-effector pose. Unlike the previously mentioned approaches, RRMC is purely reactive and provides the joint velocities for the next time instant rather than for the whole trajectory.

If the robot is redundant, there are an infinite set of joint velocities which will achieve the desired end-effector velocity. 
Typically an additional constraint such as minimising the norm of joint velocity
is applied.
Alternatively, through gradient projection and exploitation of the null space, the robot can complete sub-tasks while achieving its motion.

One such subtask is maximising the manipulability of the robot \cite{mdot0}. By projecting the manipulability Jacobian into the null space of the differential kinematic equation, the manipulator will perform the straight-line end-effector motion towards the goal pose, but also choose joint velocities such that the robot manipulability is maximised at each step. This is beneficial for the robustness of the robot operation as it decreases the likelihood the robot will approach a singular configuration. While there are numerous measures for kinematic sensitivity, the Yoshikawa manipulability index \cite{manip} is the most widely accepted and used measure for kinematic manipulability.
The manipulator Jacobian contains rows relating to translational and rotational velocities. Additionally, manipulators comprising mixed prismatic and revolute actuators have non-homogeneous Jacobians due to the different units used. 
Therefore, when using Jacobian-based manipulability measures such as \cite{manip} care must be taken to consider issues including length scaling and mixed units~\cite{manip2}.
In a dynamic environment, manipulability optimisation enables the robot to achieve
arbitrary velocity at the next time step in order to meet the needs of unpredictable scene dynamics.

RRMC can be reformulated as a quadratic programming (QP) optimisation problem. QPs provide the ability to consider constraints and costs which provide additional functionality. The work in \cite{qp0} incorporated the physical joint limits of a mobile manipulator into a QP. Recent work in \cite{mmc} introduced slack into the typical straight-line approach of RRMC -- effectively introducing extra redundancy to the problem which is especially useful for robots with six or fewer degrees-of-freedom. This work then exploited the extra redundancy to maximise the manipulability and avoid physical joint limits of manipulators through a reactive QP controller. The velocity damper approach described in \cite{pp0} was used to avoid joint limits but could equally be used to avoid obstacles.

A classical and alternate approach to reactive motion control is through potential fields where the robot is repelled from joint limits and obstacles \cite{re0, pot}. This guides the tool on a collision-free path to the goal but does not explicitly maximise the manipulability or conditioning of the robot.

Reactive control is essential for important sensor-based motion robotics problems such as visual servoing \cite{peter} or closed-loop visual grasping \cite{doug}. Both determine the end-effector velocity for the next control loop iteration, and this is challenging to achieve with planning-based methods.

This paper proposes a novel real-time motion controller which can avoid moving obstacles (for the whole robot, not just the end-effector), avoid physical joint limits and maximises the manipulability of a manipulator in a purely reactive manner.

In Section \ref{sec:qp} we incorporate the differential kinematics and manipulability maximisation into a general QP problem. We extend this to handle obstacles in Section \ref{sec:obs}, then velocity dampers and  slack variables in Section \ref{sec:dam}. We present the NEO controller in Section \ref{sec:con} and our
experimental methodology in Section \ref{sec:exp}. We conclude in Section \ref{sec:res} with a discussion of our experimental results and insights.

\section{Quadratic Programming} \label{sec:qp}

The generic form of a QP is \cite{opt0}
\begin{align}
    \min_x \quad f_o(\vec{x}) &= \frac{1}{2} \vec{x}^\top \mat{Q} \vec{x}+ \vec{c}^\top \vec{x}, \\ 
    \mbox{subject to} \quad \mat{A}_1 \vec{x} &= \vec{b}_1, \nonumber \\
	\mat{A}_2 \vec{x} &\leq \vec{b}_2, \nonumber \\
    \vec{d} \leq & \ \vec{x} \leq \vec{e}. \nonumber 
\end{align}
where $f_o$ is the objective function to minimise; $\mat{A}_1$, $\vec{b}_1$ define the equality constraints; $\mat{A}_2$, $\vec{b}_2$ define the inequality constraints, and $\vec{d}$ and $\vec{e}$ define the lower and upper bounds of the optimisation variable $\vec{x}$. 
When the matrix $\mat{Q}$ is positive definite the QP is strictly convex  \cite{opt0}. 

\subsection{Incorporating the Differential Kinematics into a QP}

The first-order differential kinematics for a serial-link manipulator are described by
\begin{equation}  \label{eq:rrmc}
    \vec{\nu}(t) = \matfn{J}{\vec{q}} \dvec{q}(t) 
\end{equation}
where $
\vec{\nu}(t) = 
\begin{pmatrix}
	v_x & v_y & v_z & \omega_x & \omega_y & \omega_z
\end{pmatrix}^\top \in \mathbb{R}^6
$
is the end-effector spatial velocity and $\matfn{J}{\vec{q}} \in \mathbb{R}^{6 \times n}$ is the manipulator Jacobian which relates $\vec{\nu}$ to the joint velocities $\dvec{q}(t)$, and $n$ is the number of joints in the robot.

We can incorporate (\ref{eq:rrmc}) into a quadratic program as
\begin{align} \label{eq:qp0}
	\min_{\dvec{q}} \quad f_o(\dvec{q}) 
	&= 
    \frac{1}{2} 
    \dvec{q}^\top 
    \mat{I} 
    \dvec{q}, \\
	\mbox{subject to} \quad
	\matfn{J}{\vec{q}} \dvec{q} &= \vec{\nu},
	\nonumber \\
	\dvec{q}^- &\leq \dvec{q} \leq \dvec{q}^+ \nonumber 
\end{align}
where $\mat{I}$ is an $n \times n$ identity matrix, $\dvec{q}^-$ and $\dvec{q}^+$ represent the upper and lower joint-velocity limits, and no inequality constraints need to be defined.
If the robot has more degrees-of-freedom than necessary to reach its entire task space, the QP in (\ref{eq:qp0}) will achieve the desired end-effector velocity with the minimum joint-velocity norm. However, there are other things we can optimise for, such as manipulability.

\subsection{Incorporating Manipulability into a QP}

The manipulability measure, devised in \cite{manip}, describes the conditioning of a manipulator, or how easily the robot can achieve any arbitrary velocity. We observe that maximising the spatial translational velocity performance of the robot is more important than rotational performance. Acknowledging this, and to avoid mixing units and associated non-homogeneity
issues, we adopt a strategy similar to \cite{ks} and use only the three rows of the manipulator Jacobian that correspond to the translational velocity $\mat{J}_t(q)$. The translational manipulability measure is a scalar
\begin{equation} \label{eq:man0}
    m_t = \sqrt{ \mbox{det} \left(
        \mat{J}_t(\vec{q}) \mat{J}_t(\vec{q})^\top
    \right) }
\end{equation}
which we can consider as a measure of the distance to a kinematic singularity.
A symptom of approaching a singularity is that the required joint velocities reach impossible levels \cite{peter} due to the poorly conditioned Jacobian. For reactive control in dynamic environments the motion required at the next step is not known ahead of time, so the ability to achieve arbitrary task-space velocity
is critical as will be demonstrated in Sec. \ref{sec:dodging}.

We can use the time derivative of (\ref{eq:man0}) and incorporate it into the QP in (\ref{eq:qp0}) to maximise manipulability while still achieving the desired end-effector velocity \cite{mmc}
\begin{align} \label{eq:qp1}
	\min_{\dvec{q}} \quad f_o(\dvec{q}) 
	&= 
	\underbrace{
		\frac{1}{2} 
		\dvec{q}^\top 
		\left( 
			\lambda_q \mat{I} 
		\right) 
		\dvec{q}
	}_{\mathrm{A}}
	\underbrace{
		\frac{}{} \!
		-
		\vec{J}_m^\top\dvec{q}
	}_{\mathrm{B}}, \\
	\mbox{subject to} \quad 
	\rlap{$	\underbrace{\phantom{\mat{J} \dvec{q} = \vec{\nu}}}_{\mathrm{C}}$}
	\matfn{J}{\vec{q}} \dvec{q} &= \vec{\nu},
	\nonumber \\
	\rlap{$	\underbrace{\phantom{\dvec{q}^- \leq \dvec{q} \leq \dvec{q}^+}}_{\mathrm{D}}$}
	\dvec{q}^- &\leq \dvec{q} \leq \dvec{q}^+
	\nonumber
\end{align}
\vspace{-13pt}
\begin{align*}
	\mathrm{A} &- \text{Minimises the norm of the velocity} \nonumber \\
	\mathrm{B} &- \text{Maximises the manipulability of the robot} \nonumber \\
	\mathrm{C} &- \text{Ensures the desired end-effector velocity} \nonumber \\
	\mathrm{D} &- \text{Ensures the the robot's velocity limits are respected} \nonumber
\end{align*}
where $\lambda_q$ is the gain for velocity-norm minimisation, and $\vec{J}_m \in \mathbb{R}^n$ is the translational manipulability Jacobian which relates the rate of change of manipulability to the joint velocities
\begin{equation} \label{eq:man1}
    \vec{J}_m^\top
    =
    \begin{pmatrix}
        m_t\ \mbox{vec} \left( \mat{J}_t \mat{H_{t1}}^\top \right)^\top 
        \mbox{vec} \left( (\mat{J}_t\mat{J}_t^\top)^{-1} \right) \\
        m_t \ \mbox{vec} \left( \mat{J}_t \mat{H_{t2}}^\top \right)^\top 
        \mbox{vec} \left( (\mat{J}_t\mat{J}_t^\top)^{-1} \right) \\
        \vdots \\
        m_t \ \mbox{vec} \left( \mat{J}_t \mat{H_{tn}}^\top \right)^\top 
        \mbox{vec} \left( (\mat{J}_t\mat{J}_t^\top)^{-1} \right) \\
    \end{pmatrix}
\end{equation}
with $\vec{J}_m^\top \in \R^n$ and where the vector operation $\mbox{vec}(\cdot) : \R^{a \times b} \rightarrow \R^{ab}$ converts a matrix column-wise into a vector, and $\mat{H}_{ti} \in \R^{3 \times n}$ is the $i^{th}$ translational component of the manipulator Hessian tensor $\mat{H} \in \R^{6 \times n \times n}$ \cite{ets}.

\section{Modelling Obstacles} \label{sec:obs}

A point in 3D space can be represented as $\vec{p} \in \mathbb{R}^3$. The distance $d$ between a point $\vec{p}_r$ on a robot and a point $\vec{p}_o$ on an obstacle is
\begin{equation}
    d_{ro} = \left|\left| \vec{p}_o - \vec{p}_r \right|\right|
\end{equation}
and a unit vector $\evec{n}_{ro}$ pointing from $\vec{p}_r$ to $\vec{p}_o$ is
\begin{equation} \label{eq:d0}
    \evec{n}_{ro} = 
    \dfrac{\vec{p}_r - \vec{p}_o}
          {d_{ro}}
    = - \evec{n}_{or} \in \mathbb{R}^3.
\end{equation}

The time derivative of (\ref{eq:d0}) is
\begin{align}
    \dot{d}_{ro}
    &= 
    \dfrac{\mathrm{d}}
          {\mathrm{d}t}
    \lVert \vec{p}_o(t) - \vec{p}_r(t) \rVert \nonumber \\
    &= 
    \evec{n}_{or}^\top 
    \big(
        \dvec{p}_o(t) - \dvec{p}_r(t)
    \big)
\end{align}

We know from differential kinematics that the end-effector velocity is $\vec{\nu}(t) = \matfn{J}{\vec{q}} \dvec{q}(t)$. Furthermore, we can calculate the velocity of any point fixed to the robot by taking the translational-velocity component of a modified manipulator Jacobian that considers the point as the robot's end-effector
\begin{align} \label{eq:j0}
    \mat{J}_p  (\vec{q}_{0 \ \cdots \ k}) &=
    \Lambda
    \left(
    \frac{\partial}
         {\partial \vec{q}_{0 \ \cdots \ k}}
    \big(
    {^0\mat{T}_{\vec{q}_k}} \bullet {^{\vec{q}_k}\mat{T}_{p}}
    \big) 
    \right) \nonumber \\
    \mat{J}_p  (\hvec{q}) &=
    \Lambda
    \left(
    \frac{\partial}
         {\partial \hvec{q}}
    \big(
    {^0\mat{T}_{\vec{q}_k}} \bullet {^{\vec{q}_k}\mat{T}_{P}}
    \big)
    \right)
\end{align}
where $k$ is the index of the link to which $p_r$ is attached, 
$\mat{J}_p \in \mathbb{R}^{3 \times k}$ is the Jacobian relating the velocity of the point $\vec{p}$ to the velocities of joints 0 to $k$,
${^0\mat{T}_{\vec{q}_k}}$ describes the pose of link $k$ with respect to the base frame of the robot,
$\bullet$ represents composition, and
$\Lambda(\cdot)$ is a function that converts the partial derivative of the pose from a tensor $\mathbb{R}^{6\times k \times k} $ to the translational velocity Jacobian $ \in \mathbb{R}^{3\times k}$. 
Note that this Jacobian is only a function of the $k$ joints preceding the link to which the point is attached. We denote the set of joints which $\mat{J}_p$ depends on as $\hvec{q}$ which is of a variable size -- if the point is attached to the end-effector then $k = n$. 

\begin{figure}[t!]
    \centering
    \includegraphics[width=8cm, height=5cm]{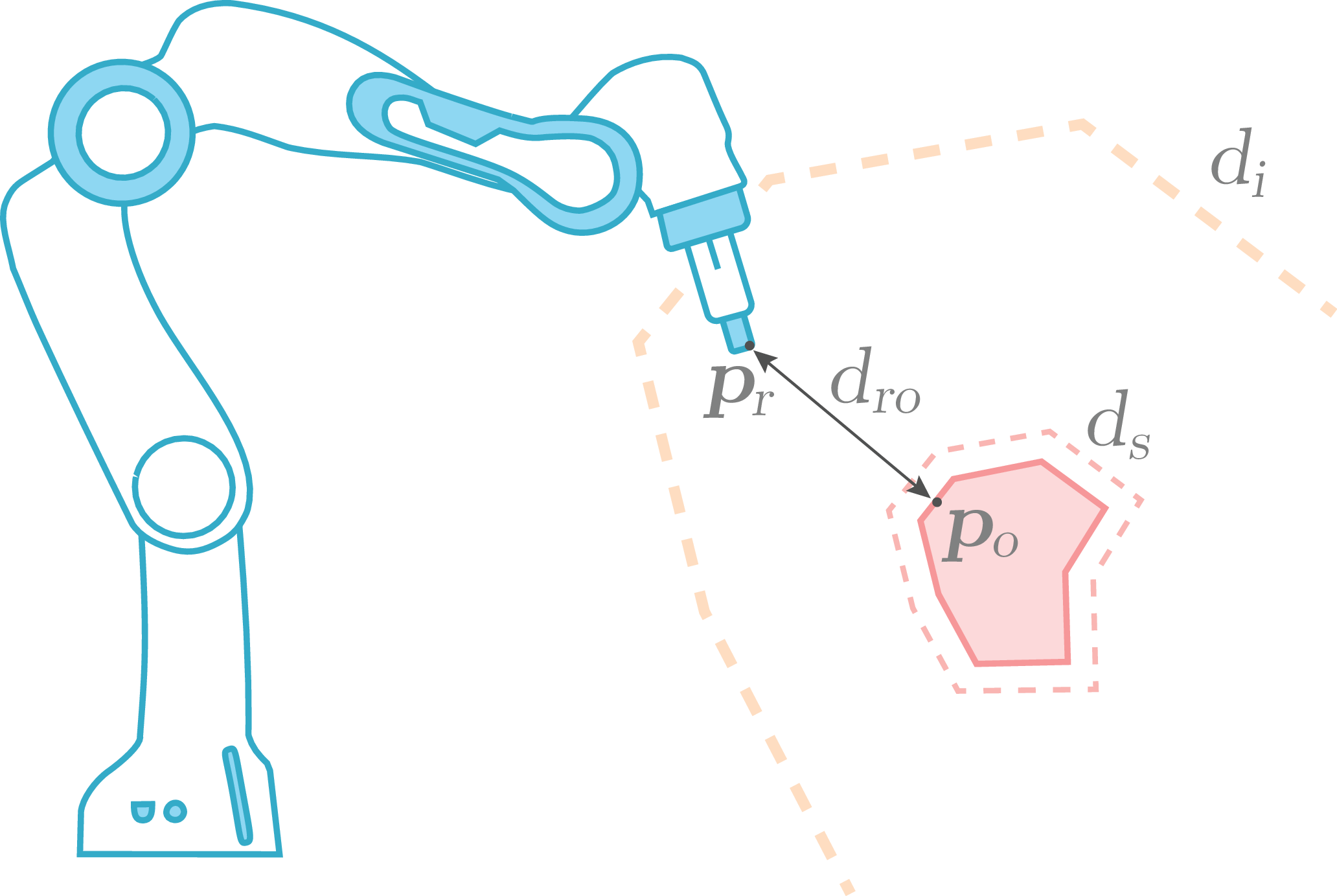}
    \caption{
        The distance $d_{ro}$ from a point on the robot $\vec{p}_r$ to a point on an obstacle $\vec{p}_o$ starts to affect the robot motion once $d_{ro}$ is less than the influence distance $d_i$. The velocity damper ensures that $d_{ro}$ is never less than the stopping distance $d_s$.
    }
    \vspace{-4pt}
    \label{fig:large}
\end{figure}

The pose ${^{\vec{q}_k}\mat{T}_{P}}$ a frame $P$ fixed to link $k$, with respect to the reference frame of link $k$, and whose origin is $\vec{p}_r$.
The orientation of this frame is arbitrary, although the direction $\evec{n}$ is expressed in this frame, so we disregard the angular velocity component from the partial derivative in (\ref{eq:j0}). More details on calculating this Jacobian can be found in \cite{ets}. Our provided software package \cite{rtb} makes it trivial to calculate these Jacobians.

Using (\ref{eq:j0}), we can now write
\begin{align}
    \dvec{p}_r(t) = \mat{J}_{p_r}(\hvec{q}) \dot{\hvec{q}}(t)
\end{align}
and the rate of change of distance becomes
\begin{align} \label{eq:d1}
    \dot{d}_{ro} (t) 
    &= 
    \evec{n}_{or}^\top
    \big(
        \dvec{p}_o(t)
        -\mat{J}_{p_r}(\hvec{q}) \dot{\hvec{q}}(t)
    \big) \nonumber \\
    &= 
    \evec{n}_{or}^\top
    \dvec{p}_o(t)
    -\evec{n}_{or}^\top
    \mat{J}_{p_r}(\hvec{q}) \dot{\hvec{q}}(t).
\end{align}

The coefficient of $\dot{\hvec{q}}(t)$ is the distance Jacobian
\begin{align} \label{eq:j1}
    \vec{J}_d (\hvec{q})
    &=
    \evec{n}_{ro}^\top
    \mat{J}_{p_r}(\hvec{q}) \in \mathbb{R}^6.
\end{align}
and substituting (\ref{eq:j1}) into (\ref{eq:d1}) we obtain the scalar
\begin{align} \label{eq:d2}
    \vec{J}_d (\hvec{q}) \dot{\hvec{q}}(t)
    &= 
    \dot{d_{ro}} (t) -
    \evec{n}_{or}^\top
    \dvec{p}_o(t)
\end{align}
which is now in a form usable for our QP.

\section{Obstacle Avoidance} \label{sec:dam}

\subsection{Velocity Dampers}

The velocity damper approach outlined in \cite{pp0} prevents robot failure by damping or restricting a velocity before limits are reached. This can be used to restrict joint motion before hitting a limit, or to constrain robot velocity before hitting an obstacle. The general velocity damper formula is
\begin{align}
    v \leq
	\xi
	\frac{d - d_s}
		 {d_i - d_s}
\end{align}
where $v$ is the rate of change of the distance $d$, $\xi$ is a positive gain which adjusts the aggressiveness of the damper, $d_i$ is the influence distance during which the damper is active, and $d_s$ is the stopping distance which  $d$ will never be less than. These distances are illustrated in Figure \ref{fig:large}. When used within an optimiser, $v$ must be found which is less than the limit set by the velocity damper.

We can incorporate obstacle avoidance into the velocity damper as
\begin{align}
    \dot{d_{ro}} (t) \leq
	\xi
	\frac{d - d_s}
		 {d_i - d_s}.
\end{align}

However, to use this as an inequality constraint of a QP we must incorporate the form of (\ref{eq:d2})
\begin{align} \label{eq:d3}
	\vec{J}_d (\hvec{q}) \dot{\hvec{q}}(t)
	\leq 
	\xi
	\frac{d - d_s}
		 {d_i - d_s} -
	\evec{n}_{or}^\top
	\dvec{p}_o(t).
\end{align}

The effect of (\ref{eq:d3}) is that the optimiser must choose joint velocities which cause the rate of change of the minimum distance  between the robot and the obstacle to increase.
Consequently, the point on the robot and the point on the obstacle will never collide while the damper is active. By adding multiple inequality constraints in the form of (\ref{eq:d3}), the robot can respond to infinitely many dynamic obstacles simultaneously. Multiple obstacles can be incorporated into the same inequality constraint by stacking instances of (\ref{eq:d3}) vertically as
\begin{align} \label{eq:d4}
	\mat{A} 
	\dvec{q}(t)
	&\leq
	\vec{b} \\
	\begin{pmatrix}
	\vec{J}_{d_0} (\hvec{q}_0) & \mathbf{0}_{1\times n - k_0} \\
	\vdots & \vdots\\
	\vec{J}_{d_l} (\hvec{q}_l) & \mathbf{0}_{1\times n - k_l} \\
	\end{pmatrix}
	\dvec{q}(t)
	&\leq
	\begin{pmatrix}
	\xi_0
	\frac{d_0 - d_s}
		 {d_i - d_s} -
	\evec{n}_{{or}_0}^\top
	\dvec{p}_{o_0}(t) \\
	\vdots \\
	\xi_l
	\frac{d_l - d_s}
		 {d_i - d_s} -
	\evec{n}_{{or}_l}^\top
	\dvec{p}_{o_l}(t)
	\end{pmatrix} \nonumber
\end{align}
for $l$ obstacles, where the Jacobians within $\mat{A} \in \mathbb{R}^{l \times n}$ have been stacked and padded with zeros to make them of constant length $n$. Obviously, given multiple obstacles and finite robot degrees-of-freedom, there will exist situations for which there is no viable solution.  

If the equality constraint in (\ref{eq:qp1}) is active, the robot will be unable to diverge from its path to dodge an obstacle and the optimiser will fail -- it is unable to satisfy both the equality and inequality constraints. To circumvent this, we will add \textit{slack} to our optimiser.

\subsection{Adding Slack}

We can augment our optimisation variable $\dvec{q}$ by adding a slack vector $\vec{\delta} \in \mathbb{R}^6$ and change the problem to
\begin{equation} \label{eq:qp2}
    \vec{\nu}(t) - \vec{\delta}(t) = \matfn{J}{\vec{q}} \dvec{q}(t)
\end{equation}
where $\vec{\delta}$ represents the difference between the desired and actual end-effector velocity -- relaxing the trajectory constraint. We reformulate our QP problem as
\begin{align} \label{eq:qp3}
    \min_x \quad f_o(\vec{x}) &= \frac{1}{2} \vec{x}^\top \mathcal{Q} \vec{x}+ \mathcal{C}^\top \vec{x}, \\ 
    \mbox{subject to} \quad \mathcal{J} \vec{x} &= \vec{\nu}, \nonumber \\
	\mathcal{A} \vec{x} &\leq \mathcal{B} \nonumber \\
	\mathcal{X}^- &\leq \vec{x} \leq \mathcal{X}^+ \nonumber
\end{align}
where
\begin{align}
	\vec{x} &= 
	\begin{pmatrix}
		\dvec{q} \\ \vec{\delta}
    \end{pmatrix} \in \mathbb{R}^{(n+6)}  \\
	\mathcal{Q} &=  \label{eq:gain0}
	\begin{pmatrix}
		\lambda_q \mat{I}_{n \times n} & \mathbf{0}_{6 \times 6} \\ \mathbf{0}_{n \times n} & \lambda_\delta \mat{I}_{6 \times 6}
	\end{pmatrix} \in \mathbb{R}^{(n+6) \times (n+6)}, \\
	\mathcal{J} &=
	\begin{pmatrix}
		\mat{J} & \mat{I}_{6 \times 6}
	\end{pmatrix} \in \mathbb{R}^{6 \times (n+6)} \\
	\mathcal{C} &= 
	\begin{pmatrix}
		\vec{J}_m \\ \mat{0}_{6 \times 1}
    \end{pmatrix} \in \mathbb{R}^{(n + 6)} \\
	\mathcal{A} &= 
	\begin{pmatrix}
		\vec{J}_{d_0} (\hvec{q}_0) & \mathbf{0}_{1\times 6 + n - k_0} \\
		\vdots & \vdots\\
		\vec{J}_{d_l} (\hvec{q}_l) & \mathbf{0}_{1\times 6 + n - k_l} \\
    \end{pmatrix} \in \mathbb{R}^{(l \times n + 6)} \\
	\mathcal{B} &= 
	\begin{pmatrix} \label{eq:gain1}
		\xi
		\frac{d_0 - d_s}
			 {d_i - d_s} -
		\evec{n}_{{or}_0}^\top
		\dvec{p}_{o_0}(t) \\
		\vdots \\
		\xi
		\frac{d_l - d_s}
			 {d_i - d_s} -
		\evec{n}_{{or}_l}^\top
		\dvec{p}_{o_l}(t)
	\end{pmatrix} \in \mathbb{R}^l \\
	\mathcal{X^{-, +}} &= 
	\begin{pmatrix}
		\dvec{q}^{-, +} \\
		\vec{\delta}^{-, +}
	\end{pmatrix} \in \mathbb{R}^{n+6}.
\end{align}

We can augment the matrices $\mathcal{A}$ and $\mathcal{B}$ with velocity dampers to ensure the controller respects joint position limits. We replace $\mathcal{A}$ and $\mathcal{B}$ with
\begin{align}
	\mathcal{A}^* &= 
	\begin{pmatrix}
		\mathcal{A} \\
		\mat{I}_{n \times n + 6} \\
	\end{pmatrix} \in \mathbb{R}^{(l + n \times n + 6)}
\end{align}
\begin{align} \label{eq:gain2}
\mathcal{B}^* &= 
	\begin{pmatrix}
		\mathcal{B} \\
		\eta
		\frac{\rho_0 - \rho_s}
			{\rho_i - \rho_s} \\
		\vdots \\
		\eta
		\frac{\rho_n - \rho_s}
			{\rho_i - \rho_s}
	\end{pmatrix} \in \mathbb{R}^{l + n}
\end{align}
where $\rho$ represents the distance to the nearest joint limit (could be an angle for a revolute joint), $\rho_i$ represents the influence distance in which to operate the damper, and $\rho_s$ represents the minimum distance for a joint to its limit. It should be noted that each row in $\mathcal{A}^*$ and $\mathcal{B}^*$ is only added if the distance, $d$ and $\rho$, is less than the corresponding influence distance, $d_i$ and $\rho_i$. When a row is added to those matrices, the velocity damper for the collision avoidance or joint limit avoidance is activated.

\section{Proposed Controller} \label{sec:con}

\begin{figure*}[t]
    \centering
    \includegraphics[width=17cm, height=5cm]{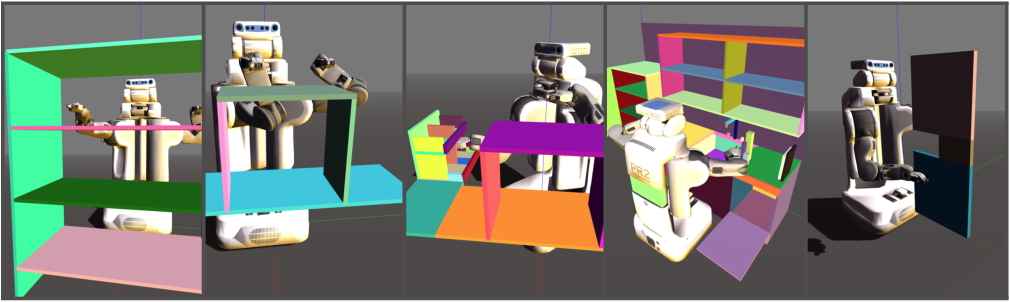}
    \caption{
        The five different test scenes in the motion planning benchmark: Bookshelf, Industrial A, Industrial B, Counter, and Tunnel.  Images are created using Swift\cite{swift} and the Robotics Toolbox for Python\cite{rtb}.
    }
    \vspace{-10pt}
    \label{fig:env}
\end{figure*}

Our proposed controller, NEO, exploits the QP described by (\ref{eq:qp3}) within a position-based servoing (PBS) scheme. This controller seeks to drive the robot's end-effector in a straight line from its current pose to a desired pose. PBS is formulated as
\begin{equation} \label{eq:pbs}
    \vec{\nu}_e = \beta \ \psi\left( (\mat[0]{T}_e)^{-1} \bullet \mat[0]{T}_{e^*} \right)
\end{equation}
where $\beta \in \mathbb{R}^+$ is a gain term, $\mat[0]{T}_e \in \SE{3}$ is the current end-effector pose in the robot's base frame, $\mat[0]{T}_{e^*} \in \SE{3}$ is the desired end-effector pose in the robot's base frame, and $\psi(\cdot) : \mathbb{R}^{4\times 4} \mapsto  \mathbb{R}^6$ is a function which converts a homogeneous transformation matrix to a spatial twist.

By setting $\vec{\nu}$ in (\ref{eq:qp3}) equal to $\vec{\nu}_e$ in (\ref{eq:pbs}) the robot will be driven towards the goal, however, not necessarily in a straight line. The robot will diverge from the path to improve manipulability, avoid joint velocity limits, avoid joint position limits, and dodge both static and dynamic obstacles. We detail the effects of changing the controller parameters in Table \ref{tab:param}.

\newcolumntype{M}[1]{>{\centering\arraybackslash}m{#1}}
\begin{table}[b!]
    \caption{Controller Parameter Description}
    \label{tab:param}

    \centering
    \renewcommand{\arraystretch}{1.3}
    \begin{tabular}{ c | m{7.2cm} }
    $\beta$ & Adjusts how fast the robot’s end-effector will travel towards the goal pose. A small value will increase the time taken to reach the goal while a large value will require large joint velocities and consequently limit the flexibility of the optimiser to achieve sub-goals. \\
    $\lambda_q$ & Adjusts the trade-off between minimising the joint velocity norm compared to maximising manipulability. A large value requires large joint velocities while a small value reduces the possible manipulability gains. \\
    $\lambda_\delta$ & Adjusts the cost of increasing the norm of the slack vector. A gain too large limits the possible additional manipulability achievable, while a gain too small leads to the possibility that the slack will cancel out the desired velocity, leaving a steady-state error.  \\
    $\xi$ & Adjusts how aggressively the robot will be repelled from obstacles. Too large results in large joint velocities, while too small will reduce obstacle avoidance performance. \\
    $\eta$ & Adjusts how aggressively the robot will be repelled from joint limits. Too large results in large joint velocities, while too small will reduce joint-limit-avoidance performance.
    \end{tabular}

\end{table}

The proposed controller is subject to local minima which can prevent it from reaching the goal pose if the direct path is blocked by an obstacle.
We coded a simple \emph{retreat} heuristic that has proven effective for many common cases: if the direct path between the end-effector and the goal is blocked by obstacles and the closest obstacle is within the influence distance $d_i$, then we bias $v_x$ and $v_y$ components of the desired end-effector velocity $\vec{\nu}_e$ to make the arm retract towards the robot's base frame.

We use our Python library, the Robotics Toolbox for Python \cite{rtb}, to implement our controller. We use the Python library $\mbox{qpsolvers}$ which implements the quadratic programming solver \cite{quad} to optimise (\ref{eq:qp3}).

\section{Experiments} \label{sec:exp}

We validate and evaluate our controller by testing on a real manipulator as well as in simulation.  In these experiments, NEO has full state information of the environment which otherwise would need to be provided by eye in hand and/or a third-person-view camera with depth sensing capabilities.

We use the following values for the controller parameters: $\xi = 1$, $d_i = 0.3\unit{m}$, and $d_s = 0.05\unit{m}$ in (\ref{eq:gain1}), $\eta = 1$, $\rho_i = 50\deg$, and $\rho_s = 2\deg$ in (\ref{eq:gain2}), $\beta = 1$ in (\ref{eq:pbs}), and $\lambda_q = 0.01$, and $\lambda_\delta = \frac{1}{e}$ in (\ref{eq:gain0}), where $e$ represents the total error between the current and desired end-effector pose. By setting $\lambda_\delta$ to be inversely proportional to pose error, the optimiser has freedom to maximise manipulability at the beginning of the trajectory while the increasing restriction ensures the end-effector continues to approach the goal. Table \ref{tab:param} describes each of the gains within NEO, as well as the consequence of making them too large or too small. In practice we found that performance was generally sound within $50\%$ to $200\%$ of the chosen gain values. As discussed in Section \ref{sec:con}, a velocity damper is incorporated into the controller when the minimum distance between the robot collision model and the obstacle is less than the influence distance $d_i$. This means there may be several velocity dampers for a single collision object, each with respect to a different robot link collision model.

For the simulated experiments, we use our open-source simulator Swift \cite{swift} with our Python library \cite{rtb} to simulate the Willow Garage PR2.
For the physical experiments, we use our Python library \cite{rtb}, ROS middleware to interface with the robot, and the 7 degree-of-freedom Franka-Emika Panda robot.

\subsection{Experiment 1: Physical Robot}

We show the merits of NEO by having it operate on a physical Panda manipulator for several different scenarios. 
The tasks reflect grasping scenarios where the end-effector starts high above a table, and the desired end-effector pose is just above the table. However, the robot must dodge one or more non-stationary obstacles present in the environment. This reflects a scenario where a robot is operating in close proximity to humans or other robots. We compare the performance of NEO in these tasks to the reactive potential field approach (PF) \cite{re0, pot}. The scenarios tested are as follows

\renewcommand{\theenumi}{\alph{enumi}}
\begin{enumerate}
    \item The robot must servo to a hypothetical grasp target on the table while a sphere of 0.05m radius travels towards the robot with a velocity of 0.2 m/s. If no action is taken, the sphere will collide with the robot's end-effector.

    \item The same as above, except there is a second sphere of 0.05m radius travelling towards the robot's elbow with a velocity of 0.2 m/s. If no action is taken, the first sphere will collide with the robots end-effector, while the second collides with the elbow.

    \item The same as above, except the desired end-effector pose changes by translating across the table with a velocity of 0.1 m/s for 4 seconds during the task. This represents a scenario where the grasp target has been disturbed. If no action is taken, the robot will collide with both spheres.
\end{enumerate}

\subsection{Experiment 2: Motion Planning Benchmark}

We compare NEO to several state-of-the-art motion planning algorithms on 198 problems over five different test scenes displayed in Figure \ref{fig:env}. The test scenes were part of a selection defined in the MoveIt! benchmark \cite{moveit, moveitb}. However, the five we selected: Bookshelves, Counter Top, Industrial A, Industrial B, and Tunnel, were the most complex.

In each problem, the robot is initialised in a defined joint configuration. The final position of the robot is defined by another joint configuration. The motion planners must plan between the initial and final configurations. However, NEO does not work by constraining the final joint configuration as it goes against the premise of task-space reactive control to know the final joint configuration. Therefore, NEO is constrained by the end-effector pose of the robot from the forward kinematics of the specified final joint configuration.

We obtained the definition of the benchmark scenes, initial and target configurations, and motion planning implementations from Trajopt's supplementary material \cite{trajopt}. We compare our results to four motion planners: Trajopt \cite{trajopt}, OMPL-RRTConnect \cite{omplr}, and OMPL-LBKPIECE \cite{omplk}, and CHOMP \cite{chomp}. As expected, the potential field approach performed extremely poorly in this experiment due to local minima, and those results have been excluded.

\section{Results} \label{sec:res}

\begin{figure*}[t!]
    \centering
    \begin{subfigure}{0.49\textwidth}
        \centering
        \includegraphics[height=5.0cm, width=8.5cm]{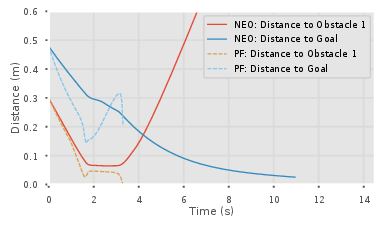}
        \vspace{-6pt}
        \caption{Experiment 1a}
        \label{fig:p1}
    \end{subfigure}
    \hfill
    \begin{subfigure}{0.49\textwidth}
        \centering
        \includegraphics[height=5.0cm, width=8.5cm]{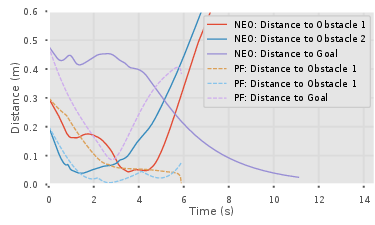}
        \vspace{-6pt}
        \caption{Experiment 1b}
        \label{fig:p2}
    \end{subfigure}
    \begin{subfigure}{0.49\textwidth}
        \centering
        \includegraphics[height=5.0cm, width=8.5cm]{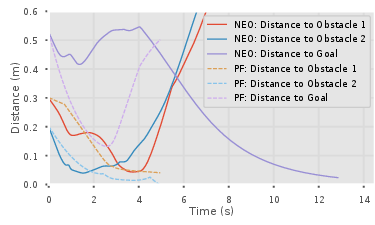}
        \vspace{-6pt}
        \caption{Experiment 1c}
        \label{fig:p3}
    \end{subfigure}
    \hfill
    \begin{subfigure}{0.49\textwidth}
        \centering
        \includegraphics[height=5.0cm, width=8.5cm]{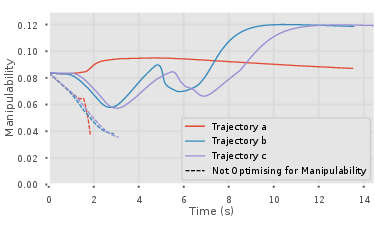}
        \vspace{-6pt}
        \caption{Experiment 1 using NEO}
        \label{fig:p4}
    \end{subfigure}
    \caption{(a-c) Euclidean distance between robot and obstacles, and robot and goal comparing NEO to the potential field (PF) method. (d) The manipulability of the robot being controlled using NEO, throughout the 3 scenarios.}
    \vspace{-16pt}
\end{figure*}

The average execution time of the NEO controller during the experiments was $9.8$\unit{ms} using an Intel i7-8700K CPU with 12 cores at 3.70GHz in a single threaded process -- a control rate just above $100$\unit{Hz}. Each collision model to obstacle pair being considered in the controller (i.e. the minimum distance between them is less than $d_i$) takes $0.5$\unit{ms}. While this is an order-of-magnitude slower than vanilla RRMC it provides significant additional capability at a rate well in excess of typical vision sensors which have a refresh rate of $30$\unit{Hz} -- our controller is much faster than the slowest part of the pipeline. Reducing the execution time is possible since many parts of the controller, such as collision checking, are parallelizable.

\subsection{Experiment 1 Results}\label{sec:dodging}
Scenario a) displayed in Figure \ref{fig:p1}, shows that for both controllers, once the obstacle gets close to the robot, the robot's end-effector stalls in its progress towards the goal. The PF controller initially repels the robot from the obstacle but eventually collides as the robot was not able to move around it. The NEO controller works correctly and dodges the obstacle while never coming closer than the stopping distance of 5\unit{cm}.

Scenario b) displayed in Figure \ref{fig:p2} is more difficult, as there are two obstacles on a collision course with the robot. The PF controller once again starts to dodge the obstacles but collides after 6 seconds. The NEO controller, while being disrupted twice from the desired trajectory, dodges both obstacles and achieves the goal pose. 

Scenario c) displayed in Figure \ref{fig:p3}, shows similar results to scenario b), where the PF controller is initially repelled from the obstacles before colliding with them and the NEO controller achieves the goal.

Figure \ref{fig:p4} displays the manipulability of the robot during each trajectory with and without manipulability optimisation. With manipulability optimisation, despite the complex nature of the robot movement to dodge dynamic obstacles and track a moving goal pose, the manipulability of the robot remains very high in all three trajectories. Without manipulability optimisation, the manipulability of the robot drops rapidly in all three scenarios. Consequently, when the robot needs large spatial velocities to dodge the dynamic obstacles, it is unable to stay within the joint velocity limits and this leads to failure. This clearly demonstrates that manipulability is an important factor for robust reactive robot performance. 

In general, controller failure is caused by the optimiser failing to find a solution that respects the joint velocity limits of the robot. Large and/or fast-moving obstacles increase the likelihood of this happening due to the large spatial velocities required to avoid collisions. This experiment shows that a low value of robot manipulability is strongly correlated with this mode of optimiser failure.

\begin{table*}[t!]
    \vspace{3pt}
    \centering
    \renewcommand{\arraystretch}{1.3}

    \caption{Experiment 2: Motion Planning Benchmark}
    \label{tab:results}

    \begin{tabular}{ c | M{2.5cm} | M{2.5cm} | M{2.5cm} | M{2.5cm} | M{2.5cm}}
    \hline
    & {NEO (ours)} & {Trajopt} & {OMPL-RRTConnect} & {OMPL-LBKPIECE} & {CHOMP-HMC} \\
    \hline\hline
    Success & 74.2\% & 81.8\% & 85.4\% & 75.8\% & 65.2\% \\
    Average time (ms) & 9.8 & 191 & 615 & 1300 & 4910 \\
    \hline
    \end{tabular}
    \vspace{-3pt}
\end{table*}

\subsection{Experiment 2 Results}

The results from Experiment 2 are displayed in Table \ref{tab:results} and show that NEO performance has a competitive success rate compared to some well-known planners.
NEO scored very well in the \emph{Bookshelf} scene, successfully completing 45 out of 45 tasks. Additionally, in the \emph{Industrial A} scene, NEO completed 61 out of 66 tasks (92\% success rate), but performed less well in the other scenes.
From Figure \ref{fig:env}, it is clear that the \emph{Bookshelf} and \emph{Industrial A} scenes contain many large broad obstacles, while \emph{Industrial B} and \emph{Counter} scenes contain many small obstacles with numerous small and enclosed spaces. NEO performs less well on this latter category.  

Our goal in this work has been reactivity not complex motion planning per se.
NEO does not perform as well as the best of the planners but for scenes which do not feature many intricate obstacles with small confined spaces it is clearly a viable alternative, and with the added benefits of reactivity such as dynamic obstacle and joint limit avoidance as well as manipulator conditioning.
NEO would ideally be employed as a local controller in conjunction with a global planner that provides the global trajectory.

Another key result from Table \ref{tab:results} is that NEO takes around 10\unit{ms} to compute the next action, while the motion planners must compute the whole trajectory before the robot can move. This ranges from 191\unit{ms} to 4910\unit{ms} for the motion planners. Consequently, they lack the reactivity required to be able to respond to dynamic obstacles, changes in the environment, or even changes in the goal pose. When testing NEO on the benchmark without optimising for manipulability, we obtained the same score of 74.2\%. This is not surprising since, unlike in Experiment 1, the obstacles are static and their influence on the control problem grows slowly as the robot moves.

\vspace{-2pt}
\section{Conclusions}
In this paper we have presented NEO, a novel Expeditious Optimisation Algorithm for manipulator motion control while dodging dynamic and static obstacles, avoiding joint limits, and maximising manipulability. Our experiments show the purely reactive and real-time NEO controller is reliable in environments containing dynamic obstacles reflective of workspaces containing multiple robots working with humans. Additionally, NEO is shown to be viable for general environments containing benches or shelves but will struggle in environments containing small confined spaces with confined spaces and is an area for future work. 
In such scenarios, NEO is better employed as a local controller when dynamic obstacles are encountered.
NEO has a strong advantage in terms of low upfront computation time.

Our approach requires the computation of the manipulability Jacobian, the manipulator Jacobian (between any points on the robot), the manipulator Hessian, and distance Jacobians.
The associated Python library, the Robotics Toolbox for Python \cite{rtb}, can conveniently compute all these values. The only required input for our library is a kinematic robot model which can be in standard or modified Denavit–Hartenberg notation, a URDF/XACRO model, or the ETS format \cite{ets}.

\bibliographystyle{IEEEtran} 
\bibliography{ref}

\end{document}